%% file: ICVGIP-Latex-Template.tex
\documentclass[sigconf]{acmart}

\usepackage{booktabs} % For formal tables
\usepackage{multirow}
\setcopyright{rightsretained}
\acmDOI{10.475/123_4}

\acmISBN{123-4567-24-567/08/06}

\acmConference[ICVGIP'18]{ICVGIP}{December 2018}{Hyderabad, India}
\acmYear{1997}
\copyrightyear{2016}

\acmPrice{15.00}

\begin{document}
\title{Unsupervised Domain Adaptation for Learning Eye Gaze from a Million Synthetic Images: An Adversarial Approach}
% \titlenote{Produces the permission block, and
%   copyright information}
% \subtitle{Initial Submission}
% \subtitlenote{The full version of the author's guide is available as
%   \texttt{acmart.pdf} document}

\author{Avisek Lahiri}
\authornote{Shares first authorship with Abhinav.}
\orcid{1234-5678-9012}
\affiliation{%
  \institution{Dept. of E\&ECE, IIT Kharagpur}
}
\email{avisek@ece.iitkgp.ernet.in}

\author{Abhinav Agarwalla}
%\authornote{The secretary disavows any knowledge of this author's actions.}
\affiliation{%
  \institution{Dept. of Mathematics, IIT Kharagpur} 
}
\email{navabhi174@gmail.com }

\author{Prabir Kumar Biswas}
\affiliation{%
  \institution{Dept. of E\&ECE, IIT Kharagpur}
  }
\email{pkb@ece.iitkgp.ernet.in}

% \author{Lawrence P. Leipuner}
% \affiliation{
%   \institution{Brookhaven Laboratories}
%   \streetaddress{P.O. Box 5000}}
% \email{lleipuner@researchlabs.org}

% \author{Sean Fogarty}
% \affiliation{%
%   \institution{NASA Ames Research Center}
%   \city{Moffett Field}
%   \state{California}
%   \postcode{94035}}
% \email{fogartys@amesres.org}

% \author{Charles Palmer}
% \affiliation{%
%   \institution{Palmer Research Laboratories}
%   \streetaddress{8600 Datapoint Drive}
%   \city{San Antonio}
%   \state{Texas}
%   \postcode{78229}}
% \email{cpalmer@prl.com}

% \author{John Smith}
% \affiliation{\institution{The Th{\o}rv{\"a}ld Group}}
% \email{jsmith@affiliation.org}

% \author{Julius P.~Kumquat}
% \affiliation{\institution{The Kumquat Consortium}}
% \email{jpkumquat@consortium.net}

% The default list of authors is too long for headers.
%\renewcommand{\shortauthors}{B. Trovato et al.}

\begin{abstract}
With contemporary advancements of graphics engines, recent trend in deep learning community is to train models on automatically annotated simulated examples and apply on real data during test time. This alleviates the burden of manual annotation. However, there is an inherent difference of  distributions between images coming from graphics engine and real world. Such domain difference deteriorates test time performances of models trained on synthetic examples. In this paper we address this issue with unsupervised adversarial feature adaptation across synthetic and real domain for the special use case of eye gaze estimation which is an  essential component for various downstream HCI tasks. We initially learn a gaze estimator on annotated synthetic samples rendered from a 3D game engine and then adapt the features of unannotated real samples via a zero-sum minmax adversarial game against a domain discriminator following the recent paradigm of generative adversarial networks. Such adversarial adaptation forces features of both domains to be indistinguishable which enables us to use regression models trained on synthetic domain to be used on real samples. On the challenging MPIIGaze real life dataset, we outperform recent fully supervised methods trained on manually annotated real samples by appreciable margins and also achieve 13\% more relative gain after adaptation compared to the current benchmark method of SimGAN \cite{apple}.
\end{abstract}



\maketitle

\input{samplebody-conf}

\bibliographystyle{ACM-Reference-Format}
\bibliography{sample-bibliography}

\end{document}

%% file: samplebody-conf.tex
%===========fig_benefit beginbs==============
\begin{figure}[!h]
\centering
\includegraphics[scale = 0.5]{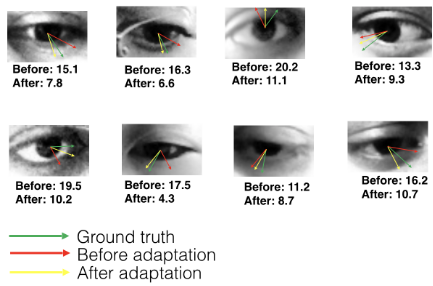} 
\caption{Examples of benefit of domain adaptation for eye gaze estimation on MPIIGaze \cite{zhang2015appearance} test samples. Predicted vectors tend to be closer to ground truth vectors after adaptation compared to vectors before adaptation (directly applying model trained on synthetic UnityEyes \cite{unity}).}
\label{fig_benefit}
\end{figure}
%===========fig_benefit ends==============
\section{Introduction}
A major reason for the contemporary success of deep learning models has been availability of large annotated datasets. It is undeniable that without abundance of labeled data, deep learning would not have reached its current pinnacle of success in numerous fields such as object recognition \cite{he2016deep, krizhevsky2012imagenet}, object detection \cite{girshick2015fast, ren2015faster, he2017mask}, action recognition \cite{wang2015action, rahmani2018learning}. Large datasets such as Imagenet \cite{russakovsky2015imagenet}, MS-COCO \cite{lin2014microsoft}, PASCAL VOC \cite{everingham2010pascal}, YouTube-8M \cite{abu2016youtube} have played a vital role in this progress. Often these datasets consists of millions of annotations which require both time and money. The question of the hour is `\textit{Can we train deep nets in smarter ways ?}' One genre of approach which is quite popular these days is to resort to automated labeled data generation from video game engines.  With rapid progress of graphics research, contemporary engines are capable of rendering high quality visual samples. For example, recent works of \cite{richter2016playing, johnson2017driving,lee2017crash} show possibility of collecting infinite amount of simulated driving scenario data from video games. Similar efforts were also seen for autonomous drones \cite{airsim2017fsr} and truck driving \cite{WinNT}. 
\begin{figure*}[!t]
\centering
\includegraphics[scale = 0.4]{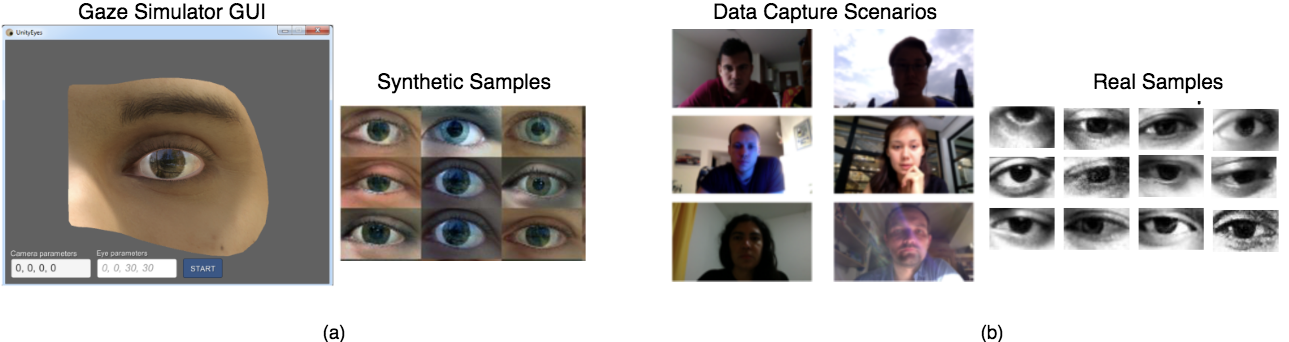} 
\caption{Visualization of two large scale gaze estimation datasets. (a): UnityEyes\cite{apple} synthetic dataset with simulator GUI and some exemplary synthetic samples; (b): MPIIGaze\cite{zhang2015appearance} dataset with typical data capture environments  and real samples. The core question in this paper is, `\textit{Can we learn a gaze estimation model from automatically annotated dataset such as UnityEyes and apply on real world dataset such as MPIIGaze with zero supervision from the latter?}'}
\label{fig_unity}
\end{figure*}
\par  While the prospect of learning from simulated data may look promising, we take a step back and ask, \textit{`Is this really a free lunch ?'} Samples from simulation engines come from a different distribution compared to real world samples. Thus discriminative models trained on synthetic data is expected to perform sub optimally on real world compared to a model which is trained solely on annotated real samples. There are two ways to tackle this problem, viz., a) improve the fidelity of graphics engine itself - this requires lot of computationally expensive optimizations and is time consuming b) project real and synthetic samples to a domain invariant representation space. In this paper, we focus on the second aspect for the particular use case of learning gaze estimation from synthetic samples generated by Unity game engine and applied on real life `in-the-wild' gaze data of MPIIGaze.
\par
\begin{figure*}
\centering
\includegraphics[scale = 0.5]{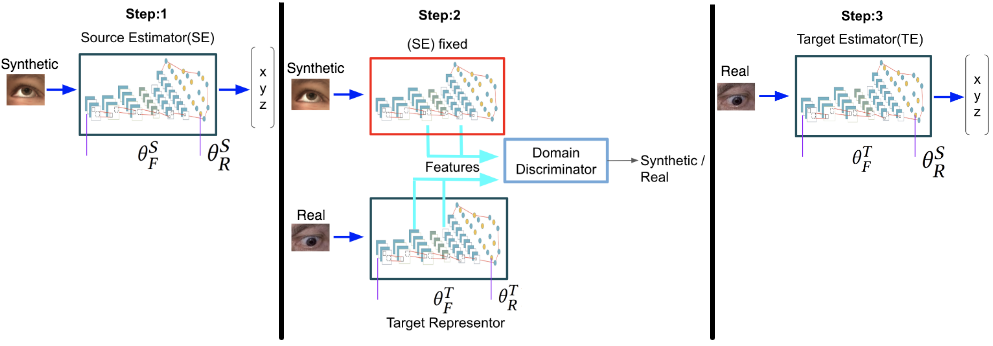} 
\caption{Stepwise flow of our model for adapting gaze estimator from automatically annotated  synthetic domain to unannotated real domain. \textbf{Step 1:} Source estimator network is trained on labels of synthetic data. Feature representation layers are denoted by $\theta_F^S$ and regression specialized layers are denoted by $\theta_{R}^S$. \textbf{Step 2:} Source estimator is frozen. Similar network, (target representer) is initialized with $SE$ with corresponding parameters, $\theta_F^T$ and $\theta_{R}^T$. Different combined layers of $\theta_F^S$ and $\theta_F^T$ are fed to a domain discriminator which distinguishes features from two domains. Target representer and domain discriminator are iteratively updated in an adversarial game paradigm \cite{goodfellow2014generative}. It is expected, that feature representations of real and synthetic samples will become indistinguishable at termination of this stage. \textbf{Step 3:} For inference on real samples, features are taken from ($\theta_F^T$) of target representer while regression specialized fully connected layers($\theta_{R}^S$) are used from source estimator for gaze estimation.}
\label{fig_flow}
\end{figure*}

 We pose the above problem as an unsupervised domain adaptation problem and leverage the recent concept of generative adversarial networks (GAN) \cite{goodfellow2014generative} to match the feature distributions of synthetic and real samples. It is a 3 stage process as depicted in Fig. \ref{fig_flow}. We perceive a deep neural as consisting of two modules, feature representer and gaze regressor.  In unsupervised domain adaptation, we assume the presence of labeled data from source domain, in our case it is the simulated/synthetic domain. We train a Source gaze Estimator (SE) on UnityEyes. In Stage2, we fix $SE$ and initialize a target representer with weights of SE. However, there are no labels available in target domain. So, intermediate features of target and source networks are fed to a adversarial domain classifier which predicts class belongingness based on features. Gradients from the domain classifier is  used for updating the target features. This step pushes the feature distribution of real samples towards synthetic samples. In Stage3, features from Target Representer are used in conjunction with regression section of Source Estimator to predict gaze on real test data. It is assumed that in Stage2, features of real and synthetic samples have become indistinguishable and thus it makes sense to use the higher order regression specific fully connected layers from source domain. We show that our model achieves 43\% relative improvement after domain adaptation compared to 30\% relative improvement achieved by the state-of-the-art method of Shrivastava \textit{et al.} \cite{apple}(SimGAN) on the challenging MPIIGaze real gaze dataset.\\
\textbf{Contributions:}
\begin{itemize}
\item This is the first demonstration of application of unsupervised(no annotation on real data) adversarial feature adaptation for 3D eye gaze estimation across simulated and real world samples
\item A data driven adaptive feature importance learning framework is introduced for assigning dynamic importance to different layers of a deep neural net for adaptation
\item Going against the usual trend of `gradient reversal' \cite{ganin2016domain} in adversarial adaptation, we empirically show that freezing source distribution prior to adaptation manifests better post adaptation performance  
\item We achieve 43\% improvement post adaptation compared to 30\% improvement by the current state-of-the-art method of SimGAN \cite{apple}
\end{itemize}
%==================================
The rest of the paper is organized as follows. In Sec. \ref{sec_related_works}, we briefly summarize some recent works on unsupervised domain adaptation and cross domain learning. Sec. \ref{sec_approach} details about our proposed approach. In Sec. \ref{sec_implementation_details}, we provide detailed description of the gaze predictor and domain discriminator networks and other related training details. Sec. \ref{sec_experiments} is related to our experimental findings and finally we conclude the paper with future scopes in Sec. \ref{sec_conclusion}.
%======== figure fig_flow starts====================
%============= figure fig_flow stops ===========
\section{Related Works}
\label{sec_related_works}
\subsection{Unsupervised domain adaptation}
Domain adaptation at feature level has been a recent genre of interest in computer vision. Closely related to our approach is the concept of Domain Adversarial Networks \cite{ganin2016domain} by Ganin \textit{et al}. to learn domain invariant features. The source network and target network share initial few layers for feature adaptation. Source network is trained on the source task while simultaneously a domain classifier discriminates two classes of features. Our approach is fundamentally different than \cite{ganin2016domain} in the sense that we initially fix the source distribution and treat it as a stationary distribution which we try to approximate with the dynamic target distribution with the adversarial training. Our approach is more aligned with the original formulation of GAN \cite{goodfellow2014generative} in which the objective of generator is to approximate a natural stationary distribution (in our case distribution of synthetic samples' features). Similar approach of \cite{ganin2016domain} was also exploited by Kamnitsas \textit{et al.} \cite{kamnitsas2017unsupervised} for  brain lesion segmentation across different datasets and it was reported that simultaneous training of source loss with domain adversarial loss requires very specific scheduling of training of each component. As shown in Fig. \ref{fig_flow}, our three stage training is very straight forward and does not require examining individual components to trigger/dampen any component of training. Ghiffary \textit{et al.}\cite{ghifary2016deep} extended DANN by replacing maximization of domain classification loss by minimization of Maximum Mean Discrepancy (MMD) metric \cite{gretton2012kernel} between features of samples from two domains.
\par Another paradigm of feature adaptation using using deep learning is to fix feature representation from both domains and then finding some subspaces to align the domains \cite{caseiro2015beyond,gopalan2011domain}. This kind of strategy was also recently applied on deep features by CORAL \cite{sun2016return} which minimized Frobenius norm between a linear projection of covariance feature matrix of source domain and target covariance matrix. 
%================================
\subsection{Learning across synthetic and real domains}
Learning from simulated/synthetic data has been an active area of research in recent times. Wood \textit{et al.}\cite{unity} used Unity game engine to generate one million synthetic eye samples to learn gaze estimator and achieved state-of-the-art performance on appearance based gaze estimation. Synthetic data coming from video games are being actively used in semantic understanding of street videos \cite{richter2016playing,johnson2017driving, lee2017crash}. This is particularly helpful because collecting street videos is tedious and sometimes impossible. For example, in \cite{lee2017crash}, the authors simulated car crashes in video games to predict in real life. These methods were particularly trained on such artificial data and had no access to real datasets. Recently, Shrivastava \textit{et al.}\cite{apple}(SimGAN) proposed an adversarial pixel domain adaptation to exploit samples from both synthetic and real domain. Their idea was to use a `pixel level refiner' network to adversarially transform annotated synthetic data coming from UnityEyes to be visually indistinguishable from real samples of MPIIGaze. A regression model trained on such transformed image dataset is expected to perform better on real samples  At the same time, a similar approach was proposed by Bousmalis \textit{et al.} \cite{bousmalis2017unsupervised} for pixel level domain adaptation with adversarial loss. We take a complementary approach to both \cite{bousmalis2017unsupervised,apple} in the sense that we adapt the feature representation of the two domains instead of pixel space. Our intuition is that, close adherence of visual properties between two domains might not necessarily indicate close performance of discriminative tasks\cite{salimans2016improved}. Thus instead of pixel space adaptation, it is more intuitive to adapt the discriminative features directly related to the task at hand.  Our approach encourages features of two domains to be similar not just based on visual appearances but also utilizes labeled data on source domain to learn task specific transferable features. This should help in gaining better relative improvement after adaptation and indeed we will see in Sec. \ref{sec_sot} that our method achieves 43\% relative improvement after adaptation compared to 30\% by \cite{apple}. 
%======================================
\section{Approach}
\label{sec_approach}
\subsection{Background on Generative Adversarial Network (GAN)}
Generative adversarial network engages two parametrized models, viz., discriminator and generator in a two-player min-max game. Realized as a feed forward neural net, the generator network takes a latent noise vector $z$ drawn from a prior noise distribution $p_z(z)$. Following \cite{goodfellow2014generative},  $z\sim \mathcal{U}[-1,1]$ (uniform distribution) and generator maps it onto an image, $y$; $G~: z~ \rightarrow y$. The other network, discriminator, has the task to discriminate samples coming from the true data distribution $p_{data}$ and the generated distribution, $p_G$. Specifically, generator and discriminator play the following game on $V(D, G)$:
\begin{equation}
\underset{G}{min}~~ \underset{D}{max}~~ V(D, G) = \mathbb{E}_{x\sim p_{data}(x)}[\log D(x)]+ ~~\mathbb{E}_{z\sim p_{z}(z)}[1 - D(G(z))]
\end{equation}
This min-max game has global optimum when $p_{data}=p_G$ and this happens when both discriminator and generator have enough capacity \cite{goodfellow2014generative}. Empirically, it has been observed that for generator, it is prudent to maximize $\log (D(G(z)))$ instead of minimizing $\log [1 - D(G(z))] $.
%================
\subsection{Unsupervised domain adaptation}
The general formulation of unsupervised domain adaptation can be stated as follows. We assume a labeled dataset, often known as source dataset, $S = (X_S, Y_S)$. In our case, $S$ is the UnityEyes dataset on which we have automated regression labels for each image.
%============
\subsubsection{\textbf{Training source regression model:}}
Using the labeled data, we can learn a parametrized source eye gaze regression function, $R_S (\cdot): ~ R_S(\theta_S(x_S)) \rightarrow \mathcal{R}^3$, where, $\theta_S(x)$ is the representation of source image $x_S$. We breakdown $\theta_S$ into two components, viz., a) feature extraction/representation section ($\theta_F^S$) and b) layers specialized for 3D gaze regression ($\theta_R^S$). Together, these parameters are grouped as $\theta_S = \{\theta_F^S, \theta_R^S\}$. Source regression network is optimized using the usual discriminative loss, $L_{reg}(X_S, Y_S)$,
\begin{equation}
 \underset{\theta_S, ~R_S}{\mathrm{argmin}} ~L_{reg}(X_S, Y_S) = ~~ \mathbb{E}_{(x_S,y_S) \sim (X_S, Y_S)} d(y_S, R_S(\theta_S(x_S))),
 \label{eq_regressor}
\end{equation}
where $d(\cdot)$ can be any distance metric. In our case we have taken Euclidean norm between the $L_2$ normalized predicted and ground truth gaze vectors.
%==================
\subsubsection{\textbf{Domain representation and adaptation:}}
For a supervised model, such as source regressor, it is usually easy to represent input images as a function of the discriminatively trained convolution layers. Different layers of the network gives different orders of task specific representations. However, due to zero annotated data, getting a corresponding representation for the target domain, $T = (X_T)$, is bit tricky. However, assuming(will touch upon this in upcoming section), we have some way of representing source and target samples with parametrized networks, $F_S$ and $F_T$ respectively, our aim will be to train a domain discriminator, $D_{\theta_D}$, a standard binary classifier, which will distinguish between $x \sim X_S$ and $x \sim X_T$ based on $F_S$ and $F_T$. Specifically, $D_{\theta_D}$ is optimized by minimizing the usual binary classification loss, $L_D$:\\
$$L_D(X_S, X_T, F_T, F_S) = -\mathbb{E}_{x_S \sim X_S}\log[D(F_S(x_S))] $$
\begin{equation}
-  \mathbb{E}_{x_T \sim X_T}\log[1 - D(F_T(x_T))]
\label{eq_disc_train}
\end{equation}
Eq. \ref{eq_disc_train} is suited for training a domain classifier with the assumption that we have finalized the domain representations, $F_S$ and $F_T$. However, domain representations needs to be optimized so as to maximize domain confusion for the discriminator. This is because, if the representations of unlabeled target domain is indistinguishable from source domain, then the regression section, $\theta_R^S$, of source domain can operate on feature section, $\theta_F^T$, of target domain for predicting 3D gaze. Thus in general, the adversarial feature adaptation optimization criteria can be written as:
$$\underset{D_{\theta_D}}{\mathrm{argmin}}~L_D(X_S, X_T, F_T, F_S)$$
$$\underset{F_S, F_T}{\mathrm{argmin}}~L_F(X_S, Y_S, D)$$
\begin{equation}
s.t~~ \gamma (F_S, F_T),
\end{equation}
where $L_F(X_S, Y_S, D)$ is feature mapping loss under the constraints of $\gamma (\cdot)$.
\par Returning back to the question of, `\textit{How to represent source and target images?}': Previous works on transfer learning and domain adaptation prefer to initialize the representation of the target domain to be exactly same as source domain but leverage different formulations of constraint, $\gamma(\cdot)$ to regularize target representation learning. Usually, $\gamma(\cdot)$ is imposed as a layerwise constraint; to be specific, a substantial number of approaches \cite{tzeng2015simultaneous,ganin2016domain} consider exact layer wise equality between the representation of the two domain. Thus, for a multi layer neural network with $L$ layers, constraint $\gamma(\cdot)^{l}$ on layer, $l$ can be expressed as:
\begin{equation}
\gamma(F_S, F_T)^l := F_S^l == F_T^l 
\end{equation}
This genre of approach is termed as `fully constrained' adaptation, wherein adaptation is performed over all the layers of representation. For a practical perspective, such layerwise equality be imposed  by weight sharing. However, fully sharing weights across two domains can lead to sub optimal performance because a single network has to handle two different domains of input.
\par To mitigate this, recent efforts focus on learning shared representations across domains. In such scenarios, $\gamma(\cdot)$ is only imposed on shared layers of the two networks. In \cite{rozantsev2018beyond}, the authors show that partial alignment of network weights leads to efficient learning for both semi supervised and unsupervised learning. Influenced by this recent trend, we also chose to adapt partial adaptation of network layers of source and target. Selection protocol for weight shared layers is described in Sec. \ref{sec_layer_selection}. 
%==============================================
\subsubsection{\textbf{Adversarial loss for feature alignment:}}
Once we decide how to represent, $F_S$ and $F_T$, and the mode of alignment (fully shared or partial), we have to decide the functional form of constraint, $\gamma(\cdot)$. A very basic approach is to impose L$_2$ loss between the shared layers \cite{tzeng2015simultaneous}. While simple from implementation point of view, recent works\cite{mathieu2015deep} have shown that L$_2$ loss is rather conservative and yields an average solution not lying on original data manifold. In our case this would mean that while adapting target features with respect to source features using L$_2$ equality constraint, the optimizer would settle for a low risk feature generation which is not viable for either source or target but expected empirical less is minimized. Such short coming can be alleviated if we leverage the adversarial loss which encourages solutions to be near to natural data manifold. Early work on gradient reversal layer \cite{ganin2016domain} propose to use to exact zero sum min-max game formulation of GAN \cite{goodfellow2014generative}, by formulating,
\begin{equation}
L_F(X_S, Y_S, D) = - L_D(X_S, X_T, F_T, F_S).
\label{eq_min_max}
\end{equation}
However, the problem with Eq. \ref{eq_min_max}, is that during initial phase of training, it is very easy for the domain discriminator to distinguish representations of the two domains and this leads to small magnitudes of gradients flowing to the target network which is trying to update itself based on these adversarial gradients. We follow, the numerical trick in \cite{goodfellow2014generative} by formulating, 
\begin{equation}
L_F(X_S, Y_S, D) = - \mathbb{E}_{x_T \sim X_T}\log[D(F_T(x_T))]
\label{eq_alternate}
\end{equation}
Eq. \ref{eq_alternate} has the same fixed point properties as that of Eq. \ref{eq_min_max} but provides higher magnitudes of gradients towards beginning of training. Note that unlike some recent adversarial adaptation approaches such as in \cite{ganin2016domain,kamnitsas2017unsupervised} where the source and target distributions are simultaneously updated, the presented approach keeps the already learnt source distribution constant and tries to align the target distribution to source. This is more in the spirit of the original GAN formulation where the objective was to approximate a stationary distribution. 
%================
\subsubsection{\textbf{Stepwise optimization}}
Combining Eqs. \ref{eq_regressor}, \ref{eq_disc_train} and \ref{eq_alternate} and noting that source representation can be stated as $F_S(x_s) = \theta_S(x_s)$, the overall optimization criteria for our entire framework can be written as:
$$\underset{F_S, R_S}{\mathrm{min}} ~L_{reg}(X_S, Y_S) = ~~ \mathbb{E}_{(x_S,y_S) \sim (X_S, Y_S)} d(y_S, R_S(F_S(x_S)))$$,
$$ \underset{D}{\mathrm{min}} ~ L_D(X_S, X_T, F_T, F_S) = -\mathbb{E}_{x_S \sim X_S}\log[D(F_S(x_S))] $$
\begin{equation}
-  \mathbb{E}_{x_T \sim X_T}\log[1 - D(F_T(x_T))]
\end{equation}
$$\&$$
$$\underset{F_S, F_T}{\mathrm{min}} ~ L_F(X_S, Y_S, D) = - \mathbb{E}_{x_T \sim X_T}\log[D(F_T(x_T))]$$
The system of equations are optimized in the following steps. To begin with, we optimize $L_{reg}(X_S, Y_S)$ independently on the labeled source domain, i.e., on UnityEyes dataset. We then fix both $R_S(\cdot)$ and $F_S$ and do not update for remaining steps of the pipeline. Since, we fix $F_S$, optimizing $L_F(X_S, Y_S, D)$ is essentially optimizing over possible alignment for $F_T$ to be indistinguishable from $F_S$. We follow the iterative optimization procedure in \cite{goodfellow2014generative} to optimize, $L_F(X_S, Y_S, D)$ and $L_D(X_S, X_T, F_T, F_S)$. Specifically, in the update step for $L_F(X_S, Y_S, D)$, the parameters of the target network adapt to align the target feature representation(coming from several shared layers as will be described in Sec. \ref{sec_layer_selection}) with source representation. In contrary, update step of $L_D(X_S, X_T, F_T, F_S)$ forces the domain classifier, $D$ to distinguish between features coming source and target domain.
%=================================
\subsubsection{\textbf{Data driven adaptive feature importance}}
Given the entire parameter set, $\theta_T$, of the target network for domain adaptation, target feature representation, $F_T(x_T)$, basically consists of concatenation of a subset of $k$ layers from the $\theta_F^T$ part of the network; $F_T(x_T) = [l_1(x_T); l_2(x_T);....l_k(x_T)]$. The naive way to adapt will be to adapt all the $k$ layers with equal importance. This has been the general trend in domain adaptation literature \cite{tzeng2017adversarial}. However, in absence of any prior, it is prudent to learn the importance of each layer from data. This can be possible by associating a learnable importance vector , $I_T \in \mathcal{R}^{k}$ which gets updated by gradients from $L_D(X_S, X_T, F_T, F_S)$ and $L_F(X_S, Y_S, D)$. Specifically, with this importance vector, modified feature representation of $x_T$ can be written as,
\begin{equation}
F_T(x_T) = F_T(x_T) \odot I_T;
\end{equation}
where, $\odot$ is the Hadamard product operator between $F_T(x_T)$ and $I_T$.
%=================================
\section{Implementation Details}
\label{sec_implementation_details}
%======================
\subsection{Network Architectures}
\label{sec_architecture}
\textbf{Gaze Estimation Network:} We use the CNN architecture for gaze estimation as reported in \cite{apple}. In Table \ref{table_gaze_architecture} we show the details of the layers of the network. Basically, the network takes input images of dimension, 35$\times$55 and is processed  by five layers of 3$\times$3 convolution with stride = 1. To ensure, invariance to local perturbations, two max pooling layers are also introduced. We consider upto layer L$_1$ as the  feature representation block ($\theta_F^S / \theta_F^T$) of the entire gaze estimator network. Next, comes the regression specialized fully connected section of the network consisting of last two fully connected layers, L$_2$ and L$_3$. Each layer throughout the network, except the last, is followed by leaky Relu non linearity with leak(negative) slope of 0.2.  The output of last fully connected layer, $FC3$, is not followed by any non linearity. It is unit normalized before we calculate the Euclidean loss between predicted and original gaze vectors.

\textbf{Discriminator Network:} The exact architecture of the discriminator depends on the genre of approach we undertake for adapting the features. For single layer adaptation, our discriminator is a 2D CNN with 3$\times$3 convolution kernel, stride = 2. This reduces the spatial resolution 2$\times$ along each dimension. First convolution has 16 channels and we double it in each layer. This is done thrice to reduce overall feature dimension by 8$\times$ along each dimension. This is followed by a fully connected layer with one output node which yields the probability of incoming features to belong to synthetic class. 
For adapting features from two layers(stacked along channel dimension), we used 3D CNNs for better exploitation of feature changes along the depth(channel) dimension. Specifically, the smaller feature maps are resized to map the resolution of the bigger maps and concatenated along channel dimension. We again follow the above principle of stagewise reduction of spatial resolution by repetitive application of 3$\times$3$\times$3 (depth, height, width) kernels with stride of 1$\times$2$\times$2. This is again followed by a fully connected layer with a single node. Leaky Relu, with negative slope of 0.2 was used after each layer, except the last layer which uses sigmoid non linearity. 
%========================
%===========================================
\begin{table}[!t]
\centering
\caption{Architecture of gaze regression network.}
\begin{tabular}{cccccc}\hline
I/P Channels & O/P Channels & Operation & Kernels & Stride & Name \\ \hline \hline
1              & 32              & Conv      & 3X3     & 1      & C$_1$         \\
32             & 32              & Conv      & 3X3     & 1      & C$_2$         \\
32             & 64              & Conv      & 3X3     & 1      & C$_3$         \\
64             & 64              & MaxPool   & 3X3     & 2      & P$_1$         \\
64             & 80              & Conv      & 3X3     & 1      & C$_4$         \\
80             & 192             & Conv      & 3X3     & 1      & C$_5$         \\
192            & 192             & MaxPool   & 2X2     & 2      & P$_2$         \\
\multicolumn{6}{l}{\hspace{30mm}Fully Connected (9600) \hspace{23mm}  FC$_1$}    \\
\multicolumn{6}{l}{\hspace{30mm}Fully Connected (1000) \hspace{23mm}  FC$_2$} \\
\multicolumn{6}{l}{\hspace{30mm}Fully Connected (3) \hspace{27mm}  FC$_3$} \\
\multicolumn{6}{c}{Unit Normalization + Euclidean Loss } \\ \hline
\label{table_gaze_architecture}
\end{tabular}
\end{table}
\section{Experiments}
\label{sec_experiments}
%================
\subsection{Training Details}
\textbf{Source gaze estimation:}
Source domain gaze estimation on UnityEyes follows usual supervised learning approach. We used Adam optimizer\cite{kingma2014adam} for  mini batch gradient descent to optimize the parameter set, $\theta_S$. Batch size was 512 and learning rate was kept at 0.001. Training was stopped when average error saturated around 2$^\circ$ on held out validation set of 10,000 samples of UnityEyes dataset.\\
\textbf{Adversarial Feature Adaptation:}
During adversarial feature adaptation, parameters ($\theta_{F}^S / \theta_{R}^S$)of source network are frozen. Parameters ($\theta_{F}^T,  \theta_{R}^T$) are initialized with their respective components from source domain. Here also, we used Adam optimizer to update target representer, ($\theta_{F}^T$) based on single or multi level adaptation. Following the iterative training procedure in \cite{goodfellow2014generative}, we update target representer in one step and the domain discriminator in next step. Learning rate was set to 0.0001 for both the competing networks. Batch size was set to 64.
It was particularly important to introduce dropout\cite{srivastava2014dropout} in the discriminator network; else the discriminator gets too powerful and the adaptation stage diverges. Specifically, we used dropout rate of 25\% for convolutional layers and 50\% for fully connected layer. 
%========================
%===============table_self_comparison begins============
\begin{table*}[!t]
\centering
\caption{Self comparison of mean angle error on MPIIGaze test set after adversarial feature adaptation(before adaptation: mean error of 14.5$^{\circ}$) for different choices of adapting feature maps. Single level adaptation represents adapting only a specific feature layer across real and synthetic domain. Double level refers to adaptation by concatenating feature maps from two different levels. C$_k$ refers to $k^{th}$ convolution layer of gaze estimator architecture. See Table \ref{table_gaze_architecture} for details of each layer.}
\begin{tabular}{llllllllll}\hline
\multicolumn{10}{c}{Single Level Adaptation}                        \\\hline\hline
     & C$_1$   &      & C$_2$  &      & C$_3$   &      & C$_4$   &      & C$_5$   \\
     & 12.7 &      & 12.8 &      & 12.5 &      & 12.1 &      & 12.0 \\\\\hline
\multicolumn{10}{c}{Double Level Adaptation}                        \\\hline\hline
C$_1$C$_2$ & C$_1$C$_3$ & C$_1$C$_4$ & C$_1$$_C5$ & C$_2$C$_3$ & C$_2$C$_4$ & C$_2$C$_5$ & C$_3$C$_4$ & C$_3$C$_5$ & C$_4$C$_5$ \\
12.6 & 12.3 & 11.9 & 10.2 & 12.4 & 11.7 & 10.7 & 12.1 & \underline{\textbf{8.8}}  & 11.8\\\hline
\end{tabular}
\label{table_self_comparison}
\end{table*}
%=====================table_self_comparison ends =============
\subsection{Dataset description}
\textbf{Unity Eyes\cite{unity}:} For source domain, we have used the automated synthetic eye gaze generation engine of UnityEyes. As shown in Fig. \ref{fig_unity}, the framework provides a graphical user interface to set up the ranges of camera and eye gaze directions. The graphics engine then randomly generates gaze samples within these ranges at 480$\times$640 resolution. Default settings were used following discussion with authors of \cite{apple}. We generated 1 million synthetic annotated examples within 7 hours. This shows the effectiveness of using graphics engines for annotated data generation. We also kept 10,000 samples as validation set. Images were center cropped to 35$\times$55.\\
\textbf{MPIIGaze\cite{zhang2015appearance}:} We used this dataset as target domain because we explicitly do not use the labels of this dataset. MPIIGaze is the largest `in-the-wild' captured eye gaze dataset consisting of data captured on consumer laptops at random everyday unconstrained environments. There are total 213,659 images from 15 participants with 80,000 samples for testing. So this dataset captures appreciable variations of real world such as different poses, lightning, indoor/outdoor, time of the day.
Images of Unity Eyes are first converted to grey scale to be compatible with MPII Gaze samples. Original range of pixel values between [0, 255] was scaled to [-1, 1] for images of both domains as a pre-processing step. 
%======================
\subsection{Pre adaptation performance}
The source regression network discussed in Sec. \ref{sec_architecture} was trained for 80,000 iterations until the mean error converged at around 1.9$^{\circ}$ on the held out validation set of UnityEyes. Before any adaptation, the source regression network incurs a mean error of 14.5$^{\circ}$ on the MPIIGaze test set. We fix the source network and make a copy as an initializer for the target network.
%============
\subsection{Selecting layers for adaptation}
\label{sec_layer_selection}
The natural question which first occurs in mind for our approach is, `\textit{Which layer(s) to adapt}?'. Choosing appropriate layers for transfer learning/domain adaptation is still an open problem, mostly studied in the context of object recognition, detection. For example, Tzeng \textit{et al.} \cite{tzeng2014deep} showed that adapting the last three fully connected layers of Alexnet gives best performance for cross domain classification. Tzeng \textit{et al.} \cite{tzeng2017adversarial}, from which our work is adopted, utilized the last fully connected layers of a classification framework for adaptation.  This makes sense for object recognition/detection because the higher order features are agnostic to local image statistics. Deeper layers are concerned for capturing global understanding for an object. However, in our case, the scenario is different. Gaze prediction requires a network to analyze local image textures yet has to manifest robustness to local perturbations. Such lower order features are mainly derived from shallower levels of the network while we need to resort to deeper channels for local invariance.  Thus there is a need to combine the best of both worlds. Our initial experiments of adapting FC$_2$ and FC$_3$ layers were not promising with post adaptation errors of 14.3$^{\circ}$ and 14.1$^{\circ}$ respectively; this shows that extreme deeper sections of fully connected layers are task specialized. Thus we keep FC$_2$ and FC$_3$ as $\theta_{R}^T/\theta_{R}^S$, while the layers till FC$_1$ are kept as $\theta_{F}^T/\theta_{F}^S$.
\par In Table \ref{table_self_comparison} we report mean error in degree on the MPIIGaze test set. Note that for every setting reported in the Table \ref{table_self_comparison} we have also adapted the FC$_1$ layer by reshaping and concatenating with the convolutional feature maps. From Table \ref{table_self_comparison} we see that adapting multiple layers yields better results compared to adapting only single layers. By adapting a combination of \{C$_3$, C$_5$, L$_1$\} we achieved lowest error of 8.8$^\circ$. Instead of the vanilla GAN loss formulation for adapting features, we also tried the Wasserstein GAN\cite{arjovsky2017wasserstein} loss formulation and it helped in reducing mean angle error to 8.2$^\circ$-a relative improvement of 43.5\% starting from 14.5$^\circ$ before adaptation.
\par For completeness of analysis, it is to be noted that we also initially experimented with adapting combinations of triple and quadruple feature maps. But the adversarial adaptation phase did not converge properly under such settings leading to negligible post adaptation improvements. Thus, moving forward, we have not included those configurations for further analysis.
%=====================================
\subsection{Comparison with `Gradient Reversal' \cite{ganin2016domain}}
Following the usual trend of applying gradient reversal technique of \cite{ganin2016domain} for domain adversarial learning, we also initially trained the source regression model and target feature alignment module(aligning combination of C$_3$C$_5$ layers)simultaneously. However, as also reported by \cite{kamnitsas2017unsupervised}, this leads to instability of training. For example, with this vanilla strategy, we could manage only up to 6$^\circ$ error on source test set while error on target domain was around 20$^\circ$. Thus neither source task nor adaptation was successful. Following, \cite{kamnitsas2017unsupervised}, we initially trained source regression model for few epochs and then pitched the adversarial learning in conjunction with source task. This culminated in getting a mean error of 3$^\circ$ on source test set and 13.3$^\circ$ on target test set. In views of absolute test set performance and ease of training the network components, our method clearly has a significant edge over gradient reversal technique.
\subsection{Comparison with state-of-the-art}
\label{sec_sot}
In Table \ref{table_results} we compare our method with recent state-of-the-art methods on MPIIGaze test set. We report results in two parts. The first part consists of methods which were trained on manually annotated gaze datasets. It is encouraging to see that our method, which has not involved any human annotation, appreciably surpasses these fully supervised methods. 
Schneider \textit{et al.}\cite{schneider2014manifold} presented a manifold alignment method for learning person independent, calibration free gaze estimation using a variety of low level features such as Local Binary Pattern(LBP), Discrete Cosine Transform(DCT) with different regression frameworks such as regression forests, Support Vector Regression (SVR) trained on Columbia gaze dataset \cite{smith2013gaze}. In Table \ref{table_results} we report their best results with SVR. In \cite{lu2014adaptive}, Lu \textit{et al.}, maps high dimensional eye features to a low dimensional gaze positions with an adaptive linear regression. The ALR helps in selecting scarce training examples via l$_1$ optimization for high fidelity gaze estimation. Sugano      \textit{et al.}\cite{sugano2014learning} created a massive 3D reconstructed  fully calibrated eye gaze dataset from head and eye pose readings from 50 subjects. The calibration includes 160 different gaze direction and 8 head poses with a total of 64,000 eye samples. Next, they learn a random forest regression model on their rendered 3D gaze models for predicting subject independent 3D gaze. Zhang \textit{et al.}\cite{zhang2015appearance} released the till date largest real life eye gaze dataset, MPIIGaze. The authors trained a multi modal deep neural network consisting of labeled information of both head pose and eye gaze.

\par In the second part we compare models which have used labels only produced by automatic rendering engines. The seminal work of Wood \textit{et al.}\cite{unity} released the UnityEyes synthetic 3D eye gaze dataset and achieves 9.9$^\circ$ error; an already improvement of 4$^\circ$ compared to best performing fully supervised method of Zhang \textit{et al.}\cite{zhang2015appearance}. As of today, SimGAN\cite{apple}, with its adversarial pixel domain adaptation across UnityEyes and MPIIGaze is the benchmark for gaze estimation on MPIIGaze. Before adaptation, SimGAN achives an error of 11.2$^\circ$ while the error goes down to 7.8$^\circ$ after adaptation - a relative improvement of 30\%.
 It is to be noted that we intentionally kept the gaze predictor network same as SimGAN \cite{apple} to get the same baseline performance before adaptation. However, SimGAN's reported pre-adaptation error of 11.2$^\circ$ was not reproducible by us with the limited information made public. Before adaptation we attain a mean error of 14.5$^\circ$ on MPIIGaze. After adaptatin, our GAN and WGAN based models achieve mean errors of 8.8$^\circ$ and 8.2$^\circ$ respectively. Thus our WGAN based framework achieves 43\% relative improvement compared to performance before adaptation; whereas SimGAN achieves a relative improvement of 30\% improvement. 
In Fig. \ref{fig_benefit} we visualize some examples showing that after adaptation the predicted gaze vectors come closer to ground truth vectors compared to the vectors before adaptation.
%=======================================
%============table_results_start=========
\begin{table}[!t]
\centering
\caption{Comparisons of mean average error (in $^\circ$) by state-of-the-art algorithms on MPIIGaze test set. Our best model achieves 8.2$^\circ$ error after adversarial feature space adaptation, a relative improvement of around 43\% compared to 14.5$^\circ$ before adaptation. Compared to us, SimGAN achieves a relative improvement of 30\% after adversarial pixel space adaptation. }
\begin{tabular}{ccr} \hline
Training Genre                                                                                                    & Method                          & Error($^\circ$) \\\hline\hline
\multirow{6}{*}{\begin{tabular}[c]{@{}c@{}}Manually Annotated\\  Real Samples\end{tabular}} & Schneider \textit{et al.}\cite{schneider2014manifold}                              & 16.5  \\
                                                                                                         & Lu \textit{et al.}\cite{lu2014adaptive}                             & 16.4  \\
                                                                                                         & Sugano      \textit{et al.}\cite{sugano2014learning}                        & 15.4  \\
                                                                                                         & Zhang \textit{et al.}\cite{zhang2015appearance}                        & 13.9  \\\hline\hline\\

\multirow{4}{*}{\begin{tabular}[c]{@{}c@{}}Auto Annotated\\  Synthetic Samples\end{tabular}}
& Wood \textit{et al.}\cite{unity}                              & 9.9   \\
& SimGAN\cite{apple} (Before Adaptation)                         & 11.2   \\
& SimGAN  (After Adaptation)                        & 7.8   \\

                                                                                                         & \textbf{Ours} (Before Adaptation)        & 14.4  \\
                                                                                                         & \textbf{Ours}(Adaptation with GAN ) & 8.8   \\
                                                                                                         & \textbf{Ours}(adaptation with WGAN) & 8.2  \\\hline
\end{tabular}
\label{table_results}
\end{table}
%=========table_results ends ========
\section{Discussion and Conclusion}
\label{sec_conclusion}
In this paper, we presented an unsupervised domain adaptation paradigm for learning to predict real life `in-the-wild' 3D eye gaze by leveraging large number of completely unannotated real gaze samples and a pool of one million automatically labeled graphics engine generated synthetic samples. Going against the traditional trend of `gradient reversal' \cite{ganin2016domain} genre of adversarial adaptation, wherein both source and target distributions are non-stationary and simultaneously updated, we chose to follow a more `\textit{GAN}' \cite{goodfellow2014generative} like approach of fixing the source distribution and trying to approximate this stationary distribution with a dynamic target distribution. Also, quite contrary to the recent approach of \cite{tzeng2017adversarial}, where the authors advocate adapting only the last layer of a deep neural net, we show that for low level and fine grained vision application such as gaze prediction, it is more prudent to adapt a multi-depth(aligning features from different depths) feature representation. Lastly, we showed that in absence of any prior assumption of importance of a layer for adaptation, it is beneficial to jointly learn the relative importance of each layer along with feature alignment. Our method achieves a very competitive absolute performance (8.2$^\circ$ post adaptation) compared to the recent benchmark of SimGAN (7.8$^\circ$ post adaptation). However, it is promising to note that our method yields a relative improvement of 43\% with respect to pre adaptation performance compared to only 30\% relative improvement by SimGAN. Our findings suggest that it might be more prudent to tackle domain adaptation in feature space compared to adaptation in absolute pixel space as done in SimGAN. 
Since our work is the first attempt of adversarial feature adaptation across Unity and MPII, an immediate extension would be combine our method and pixel adaptation approach of SimGAN. Both of these methods are complementary to each other and thus it would be an interesting approach to formulate a joint optimization of pixel and feature adaptation.
% \begin{acks}
%   The authors would like to thank Dr. Yuhua Li for providing the
%   MATLAB code of the \textit{BEPS} method.
% \end{acks}